# THOR: Thermal-guided Hand-Object Reasoning via Adaptive Vision Sampling

SOROUSH SHAHI*, Northwestern University, USA
FARZAD SHAHABI, Northwestern University, USA
RAMA NABULSI, Northwestern University, USA
GLEN FERNANDES, Northwestern University, USA
AGGELOS K KATSAGGELOS, Northwestern University, USA
NABIL ALSHURAFA, Northwestern University, USA

Wearable cameras are increasingly used as an observational and interventional tool for human behaviors by providing detailed visual data of hand-related activities. This data can be leveraged to facilitate memory recall for logging of behavior or timely interventions aimed at improving health. However, continuous processing of RGB images from these cameras consumes significant power impacting battery lifetime, generates a large volume of unnecessary video data for post-processing, raises privacy concerns, and requires substantial computational resources for real-time analysis. We introduce THOR, a real-time adaptive spatio-temporal RGB frame sampling method that leverages thermal sensing to capture hand-object patches and classify them in real-time. We use low-resolution thermal camera data to identify moments when a person switches from one hand-related activity to another, and adjust the RGB frame sampling rate by increasing it during activity transitions and reducing it during periods of sustained activity (when the system has enough information to identify the activity). Additionally, we use the thermal cues from the hand to localize the region of interest (*i.e.* the hand-object interaction) in each RGB frame, allowing the system to crop and process only the necessary part of the image for activity recognition. We develop a wearable device to validate our method through an in-the-wild study with 14 participants and over 30 activities, and further evaluate it on Ego4D (923 participants across 9 countries, totaling 3,670 hours of video). Our results show that using only 3% of the original RGB video data, our method captures all the activity segments, and achieves hand-related activity recognition F1-score (95%) comparable to using the entire RGB video (94%). Our work provides a more practical path for the longitudinal use of wearable cameras to monitor hand-related activities and health-risk behaviors in real time.

## 1 Introduction

The ubiquitous nature of mobile phones and the rise of wearables is empowering researchers to redesign the study of human behavior. Recent advances in wearable cameras are expanding the utility of these cameras to enable not only visual confirmation, but also real-time detection of human activity in real-world settings [13]. The journey of wearable cameras began to scale with the popularity of visual life logging, where wearable cameras digitally capture everyday life activities through first-person images (*i.e.*, egocentric views) [17]. Because wearable cameras gather rich data that can empower machine-learning models to accurately classify a person's interactions with objects and environments [11], this opened up its utility within the field of activity recognition, and particularly the capture of hand-related behaviors [36]. Many hand-related health-risk behaviors—such as eating, smoking, drinking, and screen use—correlate with morbidity and mortality. Traditional self-report methods impose high participant burden and are prone to under- or over-reporting errors [14, 49]. In contrast, wearable cameras have enabled a more objective method to reliably capture these hand-related activities while reducing the error and burden of recall.

Although wearable cameras hold great promise for real-world research, they face several barriers to practical implementation. Many systems record continuous video streams, resulting in redundant data and imposing heavy demands on power, storage, and computation—resources that are often scarce on embedded platforms [27, 31]. At the same time, always-on imaging raises privacy concerns that can undermine user acceptance [3]. Moreover, to adhere to the Principle of Least Privilege (POLP), data collection should be restricted to only what is strictly necessary for the task at hand. To mitigate these issues, researchers have explored low-frame-rate sampling (*e.g.*, one frame every 30 seconds) [22] or event-based photo triggering using ambient light and motion sensors [22]. While these strategies improve battery life, they risk missing fine-grained hand-object interactions that are essential for accurately capturing hand-related activities (*e.g.*, eating a quick snack, or grabbing an object). Others rely on specific gesture-based triggers [13, 45] to activate the camera during automated detection of specific movements (*e.g.*, hand-to-mouth gestures) but often fail to generalize across diverse hand-related behaviors (e.g., interacting with a smartphone), and can generate false positives in real-world settings.

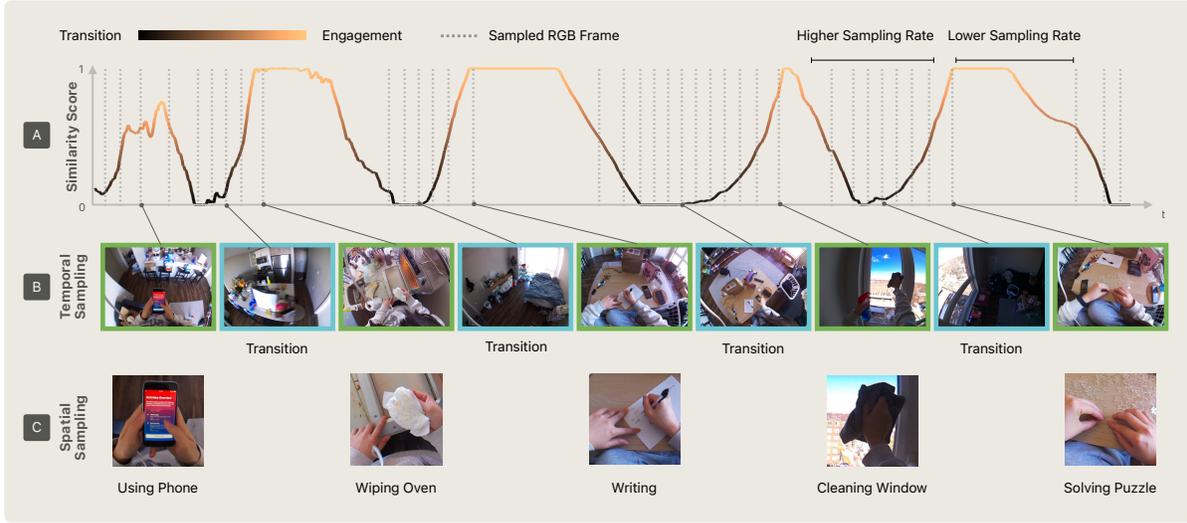

Fig. 1. Overview of THOR, a spatio-temporal RGB frame sampling method that uses thermal sensing to improve the efficiency of wearable camera systems. As illustrated in (A), the system computes a similarity score as a proxy for user activity engagement. Based on this score, it increases the temporal sampling rate during activity transitions (blue bounding boxes) and decreases it during periods of continuous engagement (green bounding boxes). The temporal sampling strategy (B) selects only the most informative frames, while the spatial sampling method (C) crops the image to focus on hand-object regions, reducing the overall volume of data to be processed.

To overcome the limitations of continuous capture, methods are needed to record only the frames that are necessary. Vision-based adaptive sampling methods [32, 50] and video summarization techniques [1, 30] address this challenge by triggering RGB video capture salient events or scene changes. For example, to save power, motion-triggered wearable cameras and context-aware smart glasses leverage environmental or behavioral cues to activate RGB video recording [12, 22, 37]. However, these solutions generally detect only coarse-grained scene-level changes (*e.g.*, transition from outdoor to indoor) or rely on external triggers, and none capture the fine-grained activity (e.g., snacking) that involve hand-object interactions that are vital for understanding health-risk behaviors.

In this work, we introduce Thermal-guided Hand-Object Reasoning (THOR), a real-time framework that leverages lower-power thermal sensing to drive adaptive RGB sampling-significantly reducing the need for continuous RGB data streams while maintaining activity recognition accuracy for both coarse- and fine-grained activity. THOR is based on two core observations: (1) *adaptive temporal sampling*: only during transitions between distinct activities is high frame-rate sampling necessary, whereas periods of sustained interaction can be captured at a much lower sampling rate; and (2) *ROI spatial sampling*: processing a cropped patch around the region of interest (ROI, around the hand and object-in-hand) can yield activity predictions on par with full-frame analysis. To our knowledge, THOR is the first egocentric vision system to employ jointly spatio-temporal adaptive sampling, reducing latency and computational demands for real-time inference, minimizing continuous RGB processing to preserve privacy, and lowering power consumption to extend device battery life (see Figure 3).

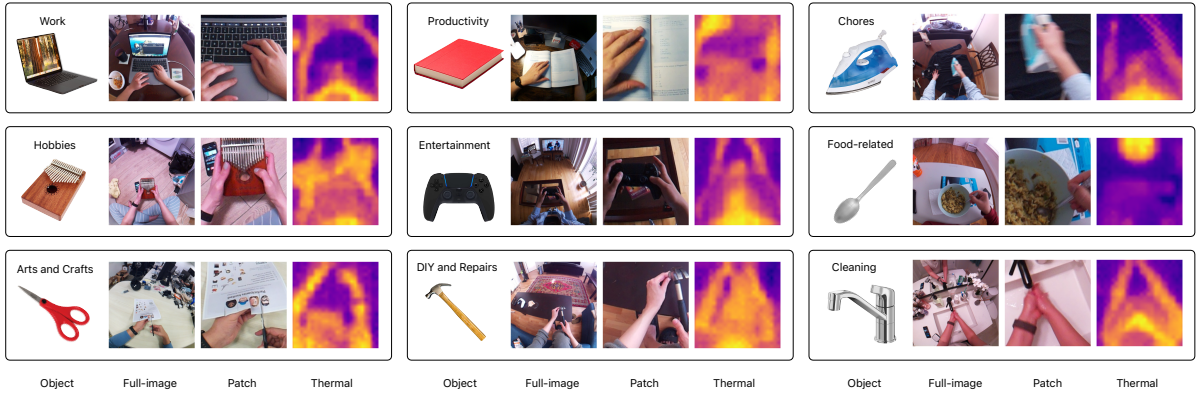

Fig. 2. Example activity categories and associated objects from our study (15 classes shown). Our method uses a thermal camera to monitor body position during these activities and samples a cropped RGB image around the hand and object-in-hand for hand-related activity recognition. In total, we collected over 30 activity classes in an in-the-wild study.

We employ a lower-power, low-resolution thermal camera to continuously capture the wearer's hand and body poses and train a contrastive learning model on 130 distinct hand-related activities, yielding embeddings that encode pose similarities and differences. By computing the cosine similarity between the current thermal embedding and a sliding window of past embeddings, we detect activity transitions when similarity drops significantly. Upon transition detection, the RGB sensor is triggered to sample at a higher frame rate, acquiring cropped patches around the hand and the object-in-hand. These patches are fed into a fine-tuned vision language model (VLM) to produce concise textual descriptions of the observed activity. We evaluate this spatio-temporal sampling pipeline on hand-related activity detection accuracy, real-time model inference latency, and device battery life, demonstrating substantial improvements in device efficiency and longevity.

In an in-the-wild study with 14 participants and over 30 activities and 2221 activity segments, our evaluation demonstrates that THOR reduces RGB sensor usage by 97%, decreases inference latency by 78%, and decreases system's overall power consumption by 48%, all while maintaining a comparable level of accuracy (95%) to state-of-the-art vision-based methods (94%) for hand-related activity recognition. To summarize, the key contributions of this paper are as follows:

- THOR I: A spatio-temporal adaptive RGB sampling method based on thermal sensing that reduces RGB sensor usage by 97% while maintaining utility (*i.e.*, high accuracy) for coarse- and fine-grained hand-related activity recognition.
- THOR II: A real-time end-to-end hand-related activity detection method from a single patch using a retrained (using both real and synthetic data), fine-tuned VLM yielding 95% F1-score with 4.51× faster inference latency compared to baseline.
- An evaluation of the proposed method in an in-wild study with 14 participants and the largest publicly available egocentric dataset, with comparisons to the state-of-the-art methods.

## 2 Background and Related Work

Recent advances in wearable sensing and egocentric vision have revolutionized hand-related activity recognition, driven by improvements in multi-modal sensor fusion, computational efficiency, and spatial understanding. The proliferation of head-mounted cameras and AR/VR devices has spurred innovations in adaptive sampling strategies to address the inherent tension between recognition accuracy and the stringent energy/compute constraints of always-on wearables. We review representative approaches across four critical dimensions: (1) adaptive frame selection, (2) multi-modal sensing architectures, (3) egocentric hand-object interaction modeling, and (4) foundation model adaptations.

### 2.1 Adaptive Frame Selection for Efficient Action Recognition

Wearable first-person cameras operate under stringent energy and compute budgets, motivating adaptive frame-sampling strategies that process video only when new information is likely. State-of-the-art methods such as AdaFrame [50] learn to skip visually redundant frames while preserving accuracy, achieving the same top-1 recognition as dense evaluation with barely eight frames per clip—over a 60% reduction in computation [50]. Complementary "any-time" cascades push the idea further: Ghodrati *et al.* add early-exit classifiers that classify easy clips with a handful of frames and reserve deeper inference for ambiguous cases, cutting per-video FLOPs by up to 70% without hurting performance [15]. Together, these works show that egocentric action cues are often concentrated in brief temporal windows, enabling aggressive sample-rate reduction.

A parallel line of research couples auxiliary sensors with event-driven capture to address both energy and privacy. Schiboni and Amft's downward-angled head camera records only the wearer's hands and torso, implicitly excluding bystanders while still capturing eating gestures [44]. Sazonov's second-generation Automatic Ingestion Monitor (AIM-2) refines this idea: an accelerometer detects hand-to-mouth motion and triggers a still image; an on-board segmentation network then redacts faces and sensitive background regions, shrinking storage and extending battery life [19]. Beyond heuristic triggers, Possas *et al.* formulate sensor selection as a reinforcement-learning problem, switching between a low-power inertial model and a vision backbone. Their policy keeps the camera active for only 8% of runtime while matching full-camera accuracy, tripling usable battery life [41].

Unlike prior work, our system employs a low-resolution thermal sensor as a continuous sentinel: thermal cues guide both *when* to sample RGB frames and *where* to crop within them. This design inherits the energy savings of sensor-triggered capture, the privacy benefits of restricting the field of view to hand-object regions, and the accuracy advantages of adaptive sampling learned in prior vision-only work. Unlike approaches driven solely by motion thresholds or fixed schedules, our system employs the thermal stream which provides real-time feedback, enabling fine-grained control over sampling rate and spatial attention precisely at moments of hand interaction.

### 2.2 Multi-Modal Sensing for Energy-Efficient Hand Activity Recognition

Wearable cameras must reconcile high recognition accuracy with the tight energy and compute budgets required for all-day use. Adaptive frame-sampling methods tackle this by evaluating only the most informative video segments. MGSampler refines this idea through motion-guided sampling that favors high-flow segments while ensuring temporal coverage, delivering both speed-ups and accuracy gains [56]. However, activities requiring static hand postures or gradual movements remain under-sampled, creating recognition gaps. Beyond heuristic sampling, Possas *et al.* model sensor scheduling as a reinforcement-learning task: their policy switches between a low-power inertial classifier and a vision backbone, sustaining accuracy with the camera active only 8% of the time [41]. However, it often optimizes short-term energy savings, leading to suboptimal camera activation sequences during prolonged activities like meal preparation. Together, these studies show that substantial portions of egocentric video can be ignored—or processed with cheaper sensors—without harming performance.

Recent work extends this efficiency further by combining auxiliary sensor "sentinels" with adaptive visual attention. Lightweight modalities such as IMUs, audio, or low-resolution thermal imagers monitor for salient motion or sound cues and trigger high-cost vision only when needed. EgoDistill fuses sparse video with head-motion IMU data to reconstruct rich embeddings, yielding two-to-three-fold FLOP reductions [47]. SmartAct adopts a thermal+RGB design to verify hand-to-mouth gestures while extending battery life by 30% [45]. Audio-visual SCSampler similarly deploys a fast audio network to pre-select salient clips, slashing computation 15× before invoking a heavier vision model [26]. These cross-modal pipelines confirm that low-power cues can "gate" camera usage, aligning processing effort with moments that matter.

Once the camera is engaged, spatial and temporal attention modules prune redundancy. Hand-centric detectors that isolate the wearer's hands and manipulated objects allow networks to crop background clutter with minimal accuracy loss [4]. Motion-guided samplers such as MGSampler continue to down-sample stagnant intervals, reserving dense inference for bursts of interaction. Privacy-aware designs fold naturally into this paradigm: ActiSight couples an RGB camera with an 8×8 thermal sensor to foreground user-specific regions while discarding bystanders [2], and HabitSense activates the RGB stream only during thermal-detected hand-to-mouth gestures, then masks the background to cut video volume by nearly 50 % while maintaining >90% F1 score[13]. Background-masking, however, inadvertently removes contextual cues critical for disambiguating similar activities-e.g., distinguishing "brushing teeth" from "drinking water" requires sink vs. kitchen countertop context.

We address the gap by using a low-resolution thermal stream as a continuous sentinel that guides *both* the temporal sampling of RGB frames and the spatial cropping toward hand-object regions. This unified, context-aware strategy inherits the energy savings of sensor-triggered capture, the privacy benefits of selective field-of-view restriction, and the accuracy advantages of adaptive vision attention—pushing egocentric monitoring closer to truly practical, all-day deployment.

## 2.3 Egocentric Hand-related Activity Recognition

Egocentric videos provide an unobstructed view of fine-grained hand–object interactions, yet first-person footage is rife with ego-motion, clutter, and rapid viewpoint shifts. Early work adapted third-person feature engineering to this regime: Ishihara *et al.* fuse local motion descriptors with global hand-shape cues, improving discrimination between visually similar grasps in cluttered scenes but still faltering under severe occlusion and fast camera shake[25]. Classical descriptors such as Histogram of Oriented Gradients (HOG), Histogram of Optical Flow (HOF), Motion Boundary Histograms (MBH), and dense trajectories capture rich kinematics, yet their frame-by-frame extraction remains compute-heavy and brittle when hands exit the field of view.

Deep networks replace hand-crafted descriptors with learned representations, but inherit new limitations. Multi-stream CNNs that add a dedicated hand-mask branch boost cooking-action accuracy by leveraging the frequent visibility of hands in egocentric footage [34]; however, they assume full-rate RGB and struggle when tools appear similar across tasks. Transformer-based pipelines such as IS-CLIP embed IMU motion patterns into visual tokens to reduce domain shift and enable zero-shot transfer [20], yet still require dense sampling and powerful GPUs—untenable on wearable hardware.

High-quality hand pose remains a prerequisite for fine-grained reasoning. Mueller *et al.* achieve real-time tracking from egocentric RGB-D by dividing localization and 3-D regression into separate CNN stages [35], but depth cameras are power-hungry and the method degrades with heavy object occlusion. AssemblyHands contributes three million egocentric frames with 3-D annotations, driving progress in single-view pose estimation and showing that stronger pose directly lifts recognition performance [38]; yet current networks still mis-predict when fingers are fully hidden behind objects.

Temporal segmentation remains a key challenge. Zhang *et al.* introduce per-pixel contact masks and context-aware augmentations to improve cross-scene generalization in fine-grained hand interactions [54], while hand-hygiene pipelines first localize candidate windows using lightweight motion cues before applying two-stream CNN classifiers [57]. These methods reduce computational cost but depend on predefined boundaries or manually tuned thresholds. More recent approaches augment Slow-Fast networks with skeletal joint streams to capture spatial–temporal hand dynamics [28], yet they require high-quality upstream pose estimates and do not directly address energy-profiling constraints. While prior methods have tackled temporal segmentation and energy constraints through heuristics or multi-stage pipelines, they often depend on dense RGB data, reliable pose estimation, or hand-crafted thresholds. In contrast, our method introduces a data-efficient alternative by adaptively sampling only the most informative RGB frames based on thermal-guided transitions. By focusing on critical hand–object moments and discarding redundant footage, we significantly cut down computational cost, storage demands, and energy consumption.

## 2.4 Foundation Models for Egocentric Vision

The availability of large vision–language models such as CLIP and BLIP-2 has sparked a wave of foundation-model research tailored to first-person video. Recent efforts pre-train billions-parameter encoders on massive egocentric corpora: EgoVideo adapts a two-tower video–language architecture to 2.6 M clips from Ego4D and EPIC-Kitchens, then fine-tunes on eight EgoVis challenge tracks—requiring 16 × A100 GPUs for 100 epochs [39]. Similar large-scale pre-training underlies EgoVLP and EgoNCE++, which couple contrastive video–text losses with temporal masking to learn hand-object interaction cues [51]. While these models deliver state-of-the-art retrieval and anticipation on egocentric leaderboards, their parameter counts (0.9–2 B) and GPU footprints hamper use in consumer wearables. Even inference can exceed 40 GFLOPs $\cdot$ s$^{-1}$, far beyond the envelope of current smart-glasses SoCs.

Foundation VLMs have also been pushed toward fine-grained hand-activity reasoning tasks. HOI-Ref extends GPT-4V to refer to the object in contact with each hand, enabling explicit question–answering about ongoing manipulations [5]. Captioning systems such as EgoInstructor retrieve instructional third-person videos to enrich first-person descriptions, improving unseen-verb coverage without manual labels [52]. VQA pipelines forecast future hand–object contacts or answer temporal queries in long clips, but nearly all rely on full-resolution RGB frames sampled at 12–30 FPS, multiplying latency and memory. When applied to 10-minute lifelog videos, a single forward pass through a billion-scale encoder can exhaust 12 GB of VRAM and several Joules of energy which is untenable on embedded devices.

The gap between research prototypes and battery-constrained hardware is evident in today's camera glasses. Meta's Ray-Ban smart glasses last ≈ 4 hours in "moderate" photo mode but only ≈ 30 minutes with the on-device AI assistant active, as VLM inference throttles the 222 mAh cell [18, 33]. Snap's latest Spectacles offer just 45 minutes of AR runtime before requiring a case recharge [7]. These limits highlight the need to reduce both the number of processed RGB frames and the size of the vision backbone for real-time egocentric analytics. Emerging work on retrieval-augmented captioning, cross-modal distillation, and thermal/IMU gating hints at lighter alternatives, yet a principled framework for selectively feeding only the most salient hand-centric views to foundation models remains open. Our approach addresses this gap by pairing a thermal sentinel with aggressive RGB down-sampling, allowing hand-related activity recognition with orders-of-magnitude fewer pixels than existing egocentric vision-based pipelines.

## 3 Study Design

Most *egocentric* vision datasets rely almost exclusively on RGB data, with only a few using thermal data for niche tasks such as hand-pose estimation [10]. In contrast, we pair a low-resolution (32×24), power-efficient thermal camera with an RGB sensor to drive adaptive sampling in real time. As shown in Figure 3-A1, the thermal modality reliably captures heat signatures from hands, arms, and torso, across diverse activities, yielding clear cues for activity-transition detection. To build our temporal sampling backbone, we conducted an in-the-wild IRB-approved (REMOVED ANONYMITY) study with 14 participants donning a device that simultaneously records RGB and thermal data. Each RGB frame was synchronized with a thermal frame and annotated with one of 130 hand-related activity labels. The final dataset comprises 2,221 labeled segments and is used for both training and evaluation of our model (see Figure 2).

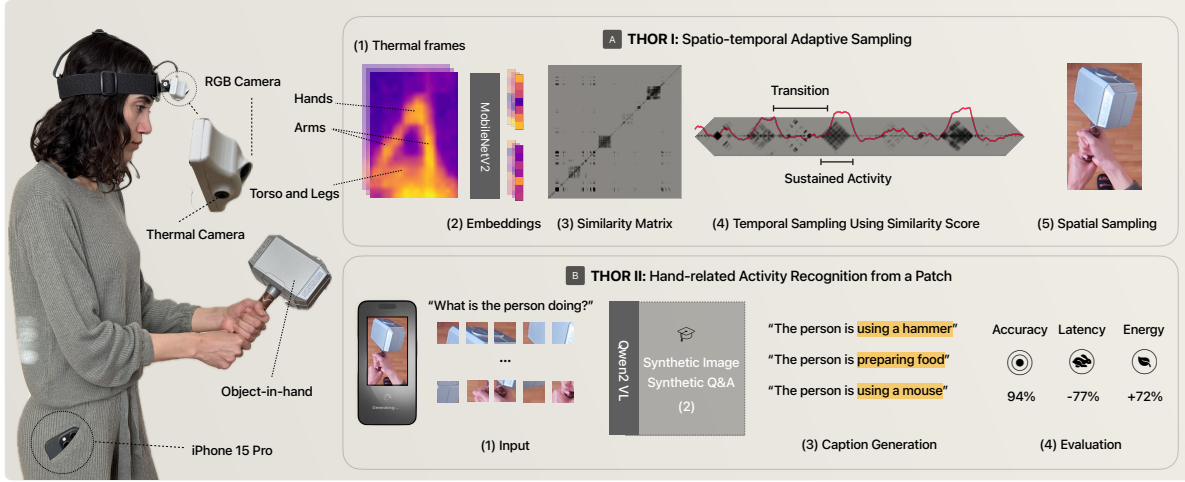

Fig. 3. Overview of the THOR system. A wearable RGB camera is augmented with a downward-facing thermal sensor to capture hand and body positions. In THOR I (A), spatio-temporal adaptive sampling is achieved by processing thermal frames (A1) through a contrastive encoder to extract embeddings (A2), which are then compared via pairwise cosine similarity (A3) to compute a similarity score (A4) that dynamically adjusts the RGB frame rate. When RGB frames are captured, a patch centered on the hand and object-in-hand (A5) is extracted and passed to THOR II (B), which performs hand-related activity recognition. THOR II takes the patch (B1) and uses synthetic data to fine-tune a vision-language model (B2), generating free-form captions describing the user's activity (B3). We evaluate THOR in terms of accuracy, latency, and energy consumption.

### 3.1 Ego4D and Synthetic Data Generation

Foundation models typically require large-scale and diverse datasets for effective training. Although pretrained models can be fine-tuned on smaller datasets, limited variability often leads to overfitting. Moreover, collecting extensive egocentric video data is both costly and time-consuming. To address these challenges, we construct a combined training set from two sources. First, we incorporate real hand-object interaction images from Ego4D [16]. Second, we augment the training data with synthetic hand-object patches generated using a pretrained diffusion model [53]. These synthetic images depict hands interacting with everyday objects sampled from the CO3D dataset [42], which provides multi-view representations of common objects. Examples of the synthesized images are provided in Appendix A.

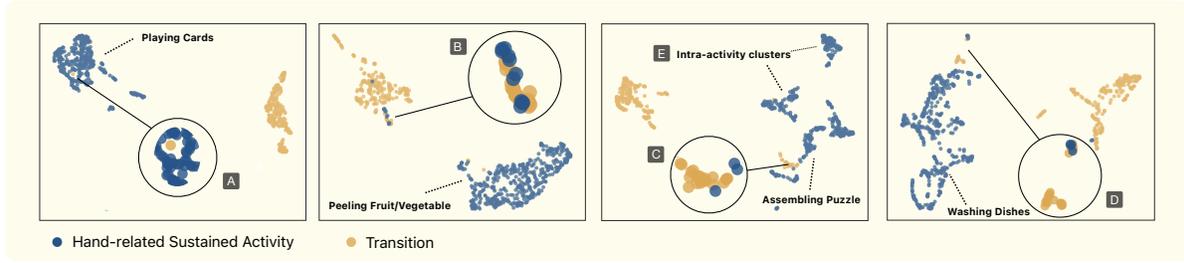

Fig. 4. In the 2D embedding space of consecutive thermal frames produced by THOR's temporal sampling method, hand-related activity frames (blue circles) form clusters that are separated from transition frames (orange circles). The regions A–D mark the moments where new activities begin. The smaller isolated clusters (E) represent subtle intra-activity movements (for example, minor posture shifts while assembling a puzzle).

### 3.2 Prototype

We combine an OV5647 RGB camera and a low-resolution MLX90614 thermal sensor, both mounted on a Raspberry Pi Zero, to capture synchronized frames at 4 frames per second. The thermal module sits just below and tilted downward relative to the RGB lens for two reasons. First, to align the thermal sensor's narrower field of view with hand- and torso-level heat signatures during head-mounted capture. Second, this orientation reduces background noise by restricting detection to the hand and object-in-hand interaction zone (see Figure 3).

### 3.3 Activity Selection

Although our sampling strategy is activity-agnostic–relying on a vision-language model rather than a fixed label set–we curated 30 representative hand-centric tasks for benchmarking, drawn from prior egocentric-vision studies [40]. We balanced the list across motion types (e.g., 'eating' vs. 'reading') and similar gestures (e.g., 'washing dishes' vs. 'washing hands'). To distribute coverage, each of our 14 participants was randomly assigned 15 activities, ensuring every task appeared multiple times across the cohort. This process produced a list of 30 daily activities, with details provided in the Appendix A.

### 3.4 Home-Based Protocol

To emulate real-world conditions, participants performed each assigned activity in their own homes for at least two minutes while wearing the head-mounted device. We captured uninterrupted RGB-thermal streams, using participants' personal appliances when possible (e.g., vacuums), or providing standard equipment otherwise. Transitions between activities were recorded continuously.

### 3.5 Free-Living Deployment

Beyond scripted sessions, we evaluated performance under true "in-the-wild" conditions. Participants wore the device for four hours of unstructured daily life, with brief five-minute removal breaks each hour to simulate real-world donning/doffing. This setup allowed us to assess algorithm robustness across natural activity patterns and device handling.

## 4 THOR I: ADAPTIVE SPATIO-TEMPORAL SAMPLING

### 4.1 Contrastive Pose Similarity Learning

To detect activity transitions, we extend contrastive learning techniques that measure image similarity between consecutive frames [8]. We adopt this idea in the thermal domain by training a model to map frames from the same activity (*i.e.*, positive pairs) close together in the embedding space, and frames from different activities (*i.e.*, negative pairs) further apart. The resulting embedding space encodes body-position variations and highlights transition points between activities (see Figure 4).

*4.1.1 Model architecture.* To support real-time, on-device processing, we use a lightweight feature extractor, MobileNetV2 [43], as the backbone of our adaptive sampling model. MobileNetV2 is pretrained on ImageNet [9], and we replace its classification head with a dense layer that outputs 64-dimensional (empirically determined) embeddings from each thermal frame. The cosine similarity between these embeddings is later used to detect transitions between activities.

*4.1.2 Batch sampling.* Effective contrastive learning hinges on careful selection of positive and negative pairs to ensure training stability [23]. We form positive pairs exclusively from frames within the same activity segment, thereby avoiding mismatches between, for example, seated versus standing portions of a single activity. Negative pairs are drawn from distinct activity classes to promote clear class separation in the embedding space. To balance training stability with training diversity, each mini-batch contains frames from 8 randomly selected activity classes, with 8 samples per class (64 frames total) sampled at each iteration.

*4.1.3 Training and validation.* We train the model using the contrastive loss function [23] on 80% of the dataset, and monitor validation accuracy on the remaining 20%. For testing, we perform leave-one-participant-out cross-validation, holding out one participant at a time to evaluate THOR I. Validation accuracy is measured using Normalized Mutual Information (NMI), which quantifies the similarity between the model's learned clusters and the ground truth activity labels, with higher values indicating better alignment and separation of activity representations. The model achieves up to 0.64 NMI on validation set across different participants. Figure 4 illustrates the embedding space learned by the model, highlighting its ability to distinguish hand-related activity frames from transition frames. The trained model is used to extract embeddings of thermal frames in the next step.

*4.1.4 Similarity score.* We extract embeddings for each frame using the model trained in the previous step, and compute cosine similarities, using the cosine distance metric, between the current frame's embedding and those from a fixed window of preceding frames (see Figure 3-A2 through A4). To mitigate noise and tailor sensitivity to the duration of different activities—ranging from short, fine-grained gestures to longer, coarse-grained episodes—we employ sliding windows of varying lengths (see Section 4.4.2). These comparisons yield a similarity matrix from which we derive the current similarity score ($S$) as the rolling-average cosine similarity over past frames. We then use $S$ to adapt the RGB camera's sampling rate in the next stage.

### 4.2 RGB Temporal Sampling

The new frame rate (FPS) of the RGB camera, is calculated using:

$$\text{FPS}_{\text{new}} = \text{FPS}_{\text{min}} + (\text{FPS}_{\text{max}} - \text{FPS}_{\text{min}}) \cdot F,$$

where $F = 1 - S$ is the effective similarity factor calculated using the similarity score in the previous step. $\text{FPS}_{\text{min}}$ and $\text{FPS}_{\text{max}}$ are two parameters that are chosen to control the number of sampled frames. To obtain $S$, we use a rolling window of size $T$ over the thermal similarity matrix. For each frame index $i$, we compute the average similarity within a submatrix defined by the window. This rolling average is then normalized using min–max normalization across the window to produce a relative similarity score. The similarity factor $S$ is defined as the inverse of this normalized score, such that higher activity transitions (*i.e.*, lower similarity) yield higher values of $F$ and therefore a higher sampling rate. The complete algorithm can be found in Algorithm 1.

---

**Algorithm 1** THOR I Dynamic FPS Update

---

**Require:** similarity matrix S, current index $i$, window $T$, previous FPS $f_{\text{prev}}$, bounds $f_{\min}$, $f_{\max}$
1: $W \leftarrow \{\max(0, i-T+1), \ldots, i\}$
2: **for** $j \in W$ **do**   ▷ rolling avg. similarity
3:     $s \leftarrow \max(0, j-T+1)$
4:     $w_j \leftarrow \text{mean}(S[s{:}j, s{:}j])$
5: **end for**
6: $r_i \leftarrow \dfrac{w_i - \min_{j \in W} w_j}{\max_{j \in W} w_j - \min_{j \in W} w_j + \varepsilon}$   ▷ min–max
7: $s \leftarrow 1 - r_i$   ▷ high change $\to$ high score
8: $f_{\text{new}} \leftarrow f_{\min} + (f_{\max} - f_{\min})\, s$
9: **return** $f_{\text{new}}$

---

### 4.3 RGB Spatial Sampling (Generating the Patch)

To limit spatial data while capturing hand-object interactions, we generate focused image patches, that represent regions of interest, guided by thermal segmentation. First, we threshold the thermal frame with Otsu's method–separating body-heat pixels from background pixels (*e.g.,* floor)–as in [2], and produce a binary mask of heat regions. We then expand this mask to cover the relevant hand-object area. There are two main cases (either the single-hand use case, or dual-hand use case). In the single-hand use case, if the mask lies to the right of the image center, we extend it 20 pixels upward and leftward; if left of the image center, we extend 20 pixels upward and rightward. In the dual-hand use case, if the mask spans both sides of the image center, we extend 20 pixels in all horizontal directions and upward. Finally, we crop the RGB frame to this expanded region and feed the resulting patch into THOR II. This targeted spatial sampling dramatically reduces storage and compute costs while preserving essential visual cues for activity recognition.

### 4.4 Definitions

*4.4.1 Activity Segment Bins.* Our dataset includes a range of activities with varying segment lengths. Certain applications, such as activity spotting of individual feeding gestures, require the capture of short activity segments, whereas others, such as counting the number of eating episodes, involve capture of longer segments. To categorize activities based on their temporal characteristics, we adopt a data-driven approach using a head-tail classification method. This technique computes two length-based thresholds, allowing us to bin segments into three categories: short (under one minute), medium (between 1 and 2.7 minutes), and long (over 2.7 minutes) segments.

*4.4.2 THOR Variants.* We introduce three variants of our method to accommodate different segment lengths: **THOR-High**, designed for high pixel sampling to capture short segments; **THOR-Mid**, targeting medium-length segments with moderate sampling; and **THOR-Low**, which minimizes pixel usage and is optimized for capturing longer activity sequences. These variants are defined by adjusting the window parameter $T$ and the maximum frame rate $\text{FPS}_{\max}$, as described in Algorithm 1.

## 4.5 Evaluation

*4.5.1 Sampling Performance.* We evaluate the sampling efficiency of THOR I by computing the ratio of pixels retained after our dynamic filtering to the total number of pixels that would have been captured by a fixed-rate, full-frame baseline and other baseline methods. Additionally, we assess coverage by calculating the proportion of activity segments for which THOR I captures at least four frames–ensuring sufficient temporal context for video-based recognition models to account for potential blur in some frames.

*4.5.2 Baseline Methods.* We evaluate THOR I against three minimum-rate, uniform-sampling baseline methods, each tuned to guarantee at least four frames per activity segment across three duration categories (short, medium, long). For each category, we select the lowest constant sampling rate meeting this requirement: 1) Uni-Low: one frame every 17 second (to capture mainly long segments); 2) Uni-Mid: one frame every 8 seconds (to capture medium and long segments); and 3) Uni-High: one frame every 2 seconds (to capture all segments). These baselines enable a fair comparison of data-capture efficiency independent of the camera's native frame rate.

*4.5.3 Power Consumption.* We measured the power consumption of key components in our system, specifically the RGB camera, thermal camera, on-device machine learning model, and the network transmission of selected patches. To capture accurate power usage, we utilized a USB power meter that directly measured the current drawn by these components during operation. These measurements were then compared to the power consumption of a baseline RGB wearable camera system, which continuously streams full-resolution images to a mobile device for processing, to highlight the efficiency improvements achieved through our method.

## 5 THOR II: Hand-related Activity Recognition from a Patch

### 5.1 Overview

The second part of our method, focuses on recognizing hand-related activities using the patches extracted from THOR I. Previous vision-based approaches for egocentric video understanding, such as TimeSformer [6] and Swin Transformers [29], often experience performance degradation when inputs are limited to a few frames or confined to small regions, such as hand-object patches. To address this limitation, we leverage recent advances in multi-modal foundation models, specifically vision-language models (VLMs), for visual recognition. Unlike conventional classification models, VLMs can generate captions for egocentric images using free-form vocabulary prompts (*i.e.*, open-set classification) enabling the recognition of a broader range of hand-related activities.

THOR II introduces techniques to enhance the utility of patches for automatic hand-related activity recognition and compares the resulting performance to prior deep learning methods. While foundation models are typically large and resource-intensive, we fine-tune a relatively lightweight VLM [48] (approximately two billion parameters) to generate concise captions focused on hand-related activities from patch inputs. This strategy yields two key benefits, illustrated in Figure 5: (1) using patches instead of full-frame images reduces the number of input tokens processed by the model, and (2) fine-tuning the model to produce short, targeted captions reduces the number of output tokens. Both factors contribute to significant reductions in latency and power consumption, as detailed in Section 6.2.5. In the following subsections, we describe the fine-tuning methodology and evaluation setup in detail.

### 5.2 Synthetic-Augmented Teacher–Student Method

*5.2.1 Teacher–Student Framework and Efficient Fine-tuning.* For each synthetic image, we prompt a large-scale VLM (72 billion parameters) to generate three *short* question–answer pairs describing the hand-object interaction. The generated responses are post-processed and used as supervision for fine-tuning a smaller model. Examples of the synthesized image-text pairs are provided in Appendix A. This teacher–student framework enables knowledge transfer from the large model to the smaller one with minimal human annotation, allowing us to use the entire dataset collected from the study as a test set. Since the fine-tuning operates on cropped patches and concise captions, both the number of input and output tokens are substantially reduced, resulting in lower computational and power demands. This design makes the approach practical for efficient real-time applications.

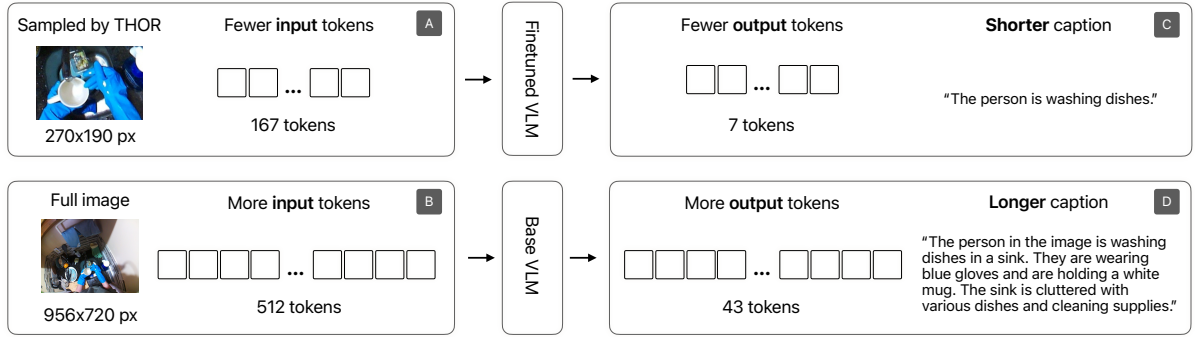

Fig. 5. By using a low-resolution patch as input, the model processes fewer visual tokens compared to full-image input (A–B). In addition, fine-tuning the model to generate concise textual descriptions results in fewer language tokens (C-D), leading to reduced latency and lower power consumption on the device.

*5.2.2 Model Architecture.* Our objective is to develop an efficient model that maintains high recognition accuracy while remaining suitable for resource-constrained devices. Although the performance of VLMs generally improves with increasing model size, larger models typically require powerful hardware such as high-memory GPUs. Given that our target platform is an iPhone, we employ a relatively lightweight model, Qwen-2 VL with 2 billion parameters [48], to generate activity captions. This choice strikes a balance between accuracy and computational efficiency, enabling practical on-device execution.

*5.2.3 Training details.* Training VLMs typically requires significant computational resources, including high-end GPUs. To make fine-tuning more efficient, we use Low-Rank Adaptation (LoRA) [24], which reduces the number of trainable parameters by injecting low-rank update matrices into the model. We set the rank to 8, with an alpha value of 8 and a dropout rate of 0.05. This approach speeds up training, without negatively affecting validation accuracy. To prevent overfitting, we reserve 30% of the data for validation and apply early stopping based on validation performance.

## 5.3 Evaluation

*5.3.1 Hand-related Activity Recognition.* Since the captions generated by the vision-language model are open-ended, the model is not restricted to a fixed set of predefined activities. However, for evaluation purposes, we assess performance based on 30 primary activities included in our study. One challenge is that an activity can be described many ways, for example, "using a phone" could be described as "texting," or "vacuuming" could be phrased as "cleaning the floor." To handle this variability, we manually construct a bag of keywords for each ground-truth activity. These keywords include synonymous actions or tasks commonly associated with the activity. For instance, the activity "cutting food" includes keywords such as "cutting potato" and "cutting onion," while "using a laptop" may include "working on a laptop." A complete list of activity-keyword mappings is provided in the Appendix A. We evaluate model performance by matching generated captions to the corresponding keyword sets and report both F1-score and accuracy averaged across both participants and across activity classes.

|        | P1   | P2   | P3   | P4   | P5   | P6   | P7   | P8   | P9   | P10  | P11  | P12  | P13  | P14  | **Mean** |
|--------|------|------|------|------|------|------|------|------|------|------|------|------|------|------|----------|
| THOR-H | 3.42 | 2.79 | 2.32 | 3.45 | 1.87 | 3.87 | 3.42 | 3.37 | 3.67 | 4.39 | 2.44 | 4.45 | 2.15 | 1.31 | **3.07** |
| THOR-M | 1.19 | 0.97 | 0.83 | 1.24 | 0.65 | 1.43 | 1.20 | 1.23 | 1.30 | 1.58 | 0.86 | 1.56 | 0.76 | 0.49 | **1.09** |
| THOR-L | 0.29 | 0.23 | 0.20 | 0.29 | 0.16 | 0.33 | 0.29 | 0.29 | 0.31 | 0.38 | 0.20 | 0.37 | 0.18 | 0.12 | **0.26** |

Table 1. Percentage of data utilized per participant for each THOR configuration.

*5.3.2 Baseline Methods.* We compare our method against baseline approaches that use full video data and state-of-the-art deep learning models for hand-related activity recognition. For image-based recognition, we evaluate SwinIR [29], which provides a fair comparison since our method operates on a single image. For video-based recognition, we evaluate TimeSformer [6], a leading model for spatiotemporal video understanding. In contrast to our method (where we trained on Ego4D and synthetic data, all the baseline models are trained on our dataset and validated using leave-one-out cross validation.

*5.3.3 Generalization.* To evaluate THOR II's ability to generalize beyond simple keyword matching, we applied our method to hand-object patches from Ego4D, the largest available egocentric dataset [16]. Because Ego4D lacks thermal imagery and only annotates hand-object interactions in a subset of videos, we first deployed a hand-object detection model [46] to extract relevant patches across the full dataset. We then ran THOR II on these patches to generate captions for over 100 annotated activity scenarios. To evaluate generalization beyond keyword-based methods, we embed the generated captions and compare their cosine similarity to the embeddings of activity narrations provided by the dataset. Higher similarity indicates more accurate captions for hand-related activities. This analysis highlights THOR II 's ability to generalize across datasets and recognize activities without relying on handcrafted keyword lists.

*5.3.4 Latency.* We measure the system's inference latency, both with and without THOR II, in seconds to assess the impact of using patches and generating short descriptions on the model's inference speed. Specifically, we compare the latency when using hand-object patches and fine-tuned short captions to the latency when processing full-frame images with captions generated prior to fine-tuning.

*5.3.5 Power Consumption.* We evaluated the impact of THOR on the power consumption of the user's phone, specifically an iPhone 15 Pro with a 3274 mAh battery. Streaming full-size wearable camera images (956×720 pixels) into a VLM running on the phone in real-time could rapidly deplete the battery. However, our method reduces both the temporal and spatial amount of information processed, which in turn lowered the power required for image transfer to the phone. This decreases the computational load during VLM inference and significantly improves power efficiency. To assess this, we implemented an iOS app that receives patches sampled by THOR I and ran THOR II *on-device* in real-time. We then measure the phone's battery consumption using a USB power meter and compare the results against baselines, including the power consumption during inference using the model before fine-tuning.

## 6 Results
## 6.1 **THOR I**: Adaptive Saptio-temporal Sampling

*6.1.1 Sampling Performance.* Table 1 summarizes the proportion of data utilized by our spatio-temporal sampling approach across all participants. THOR-Low achieves complete coverage of all long activity segments using only 0.26% of the original data. By increasing data usage to 1.09% (an additional 0.83%), THOR-Mid successfully captures all medium and long segments. To achieve full segment coverage across all granularities, THOR-High requires 3.07% of the data. The variations between participants reflect differences in individual behavior patterns and activity transitions, which might be influenced by factors such as attention span. For instance, P12 exhibited frequent transitions during daily routines, resulting in the highest data usage under THOR-High (orange box in Table 1) to cover all segments. In contrast, Participant P14 engaged in prolonged stationary periods, often in social settings, leading to fewer short and medium segments and the lowest data consumption for capturing short segments (green box in Table 1).

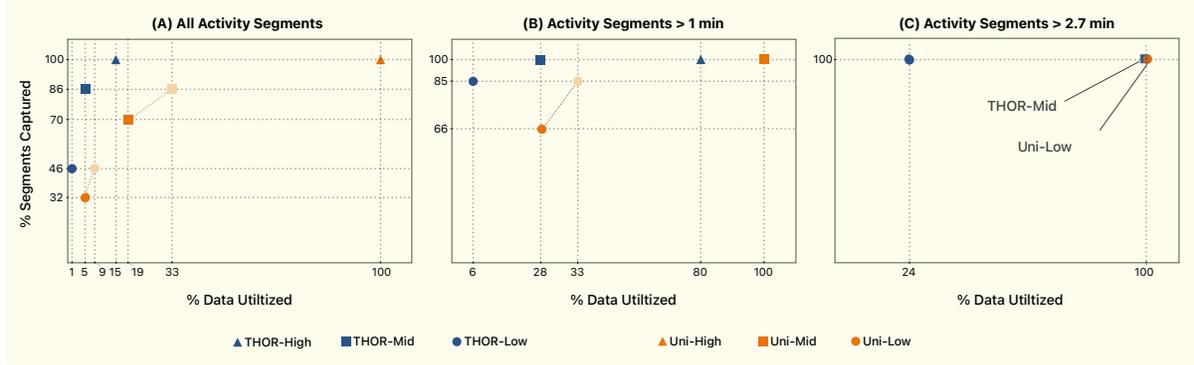

Fig. 6. Coverage of segments with different length using different variations of THOR and baselines methods.

*6.1.2 Comparison to Baselines.* Figure 6 compares our spatio-temporal sampling method with uniform baseline strategies. THOR-Low captures 100% of the long activity segments while consuming only 24% of the data used by the Uni-Low baseline (Figure 6-C). Additionally, it captures 85% of long and medium activity segments using just 6% of the data required by Uni-Mid (Figure 6-B). THOR-Mid achieves complete coverage of short and medium segments while using 28% of the data consumed by Uni-Mid (Figure 6-B). Similarly, THOR-High captures segments of all durations, including short segments, using only 15% of the data compared to Uni-High (Figure 6-A). Notably, THOR-Mid consumes approximately the same amount of data as Uni-Low, yet captures 34% more medium-length segments (Figure 6-B). All reported values are normalized with respect to their corresponding baselines. Overall, THOR-High achieves complete segment coverage while utilizing only 3% of the original (unfiltered) data. A detailed participant-level analysis of data usage is provided in Table 1.

*6.1.3 Power Consumption.* Figure 7-A illustrates the power consumption profile of individual system components under THOR I, compared to a baseline method that continuously streams frames from the RGB sensor. In the baseline setup, 75% of the power is consumed by the RGB sensor for frame capture, while 25% is used for transmitting frames to the phone for real-time processing. In contrast, our method's adaptive frame sampling significantly reduces the power consumption of both the RGB sensor (by 73.2%) and network transmission (by 22%). The addition of the thermal sensor and THOR I 's deep learning model introduces minimal overhead, resulting in a net reduction of 24% in total system power consumption relative to the baseline.

Table 2 profiles the energy cost of the core sensing pipeline under five configurations. Using the RGB sensor alone establishes a 60 mWh baseline. Adding the thermal sensor increases the consumption by 15 mWh, confirming that thermal camera itself is not the dominant drain. Incorporating both RGB and thermal sensing incurs only a marginal 5 mWh overhead. By contrast, continuously streaming RGB frames to the network is the most expensive setting (80 mWh), highlighting the impact of wireless transmission on battery life. Running our on-device deep-learning model while sampling RGB and thermal data using THOR I requires 69 mWh—15% less than off-loading the stream.

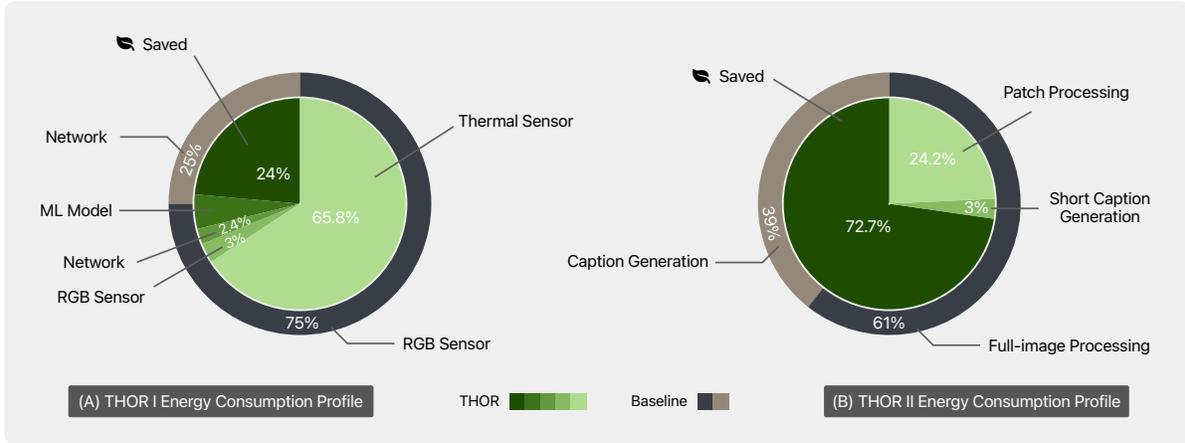

Fig. 7. Power consumption profile of different components in THOR I (A) and THOR II (B) compared to baseline methods.

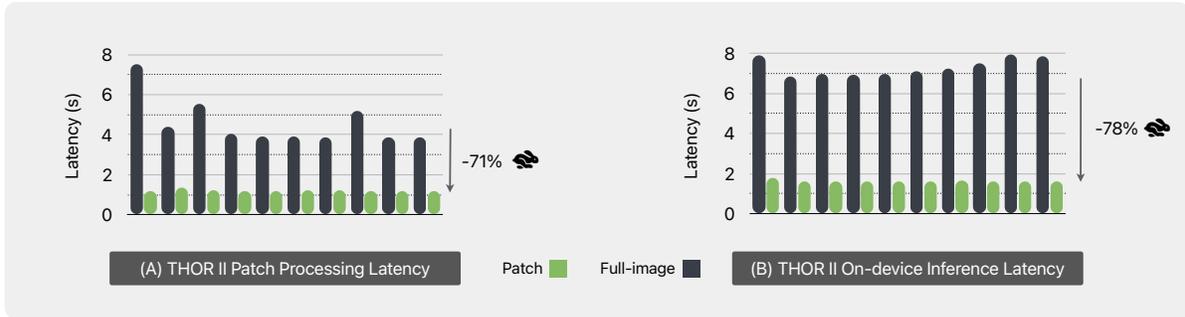

Fig. 8. Input (A) and end-to-end (B) Latency of THOR II compared to baseline method.

| Component | Current (mAh) | Power (mWh) |
|---|---|---|
| Thermal Only | 10 | 50 |
| RGB Only | 12 | 60 |
| RGB + Thermal | 13 | 65 |
| RGB + Network (Continuous Stream) | 16 | 80 |
| RGB + Thermal + Deep Learning Model | 14 | 69 |

Table 2. Power consumption of different components of the wearable device.

| Setting | Current (mAh) | Power (mWh) |
|---|---|---|
| THOR II Input Only | 6 | 31 |
| THOR II Input–Output | 7 | 36.5 |
| Baseline VLM Full-Image Input Only | 17 | 80 |
| Baseline VLM Full-Image Input–Output | 28 | 131 |

Table 3. Power consumption of iPhone 15 Pro under different settings.

## 6.2 THOR II: Hand-related Activity Recognition from a Patch

*6.2.1 Hand-related Activity Recognition.* Table 4 reports the average performance of THOR II in recognizing 30 hand-related activities across all participants. THOR II employs THOR-High as its sampling strategy, which processes only 3% of the original video data, and achieves 96% precision, 94% recall, and 95% F1-score. These results are comparable to a full-video baseline that uses Qwen2-VL, with 94% F1-score observed for the baseline. Figure 10 shows the recognition accuracy of different THOR variants across activity segment durations. THOR II achieves F1-scores of 86%, 86%, and 81% under THOR-High (which detects all activities), THOR-Mid (which detects medium and long segments), and THOR-Low (which detects long segments), respectively. Each variant is evaluated on a different subset of the entire data set based on its intended sampling policy. In summary, when averaged across all activities, THOR-High provides performance comparable to full-video inference while using only 3% of the data, with only a 2.5% reduction in F1-score and a 2% drop in accuracy.

| **THOR II**[†] | | | | Qwen2-VL[‡] | | | | SwinIR[‡] | | | | TimeSformer[‡] | | | |
|---|---|---|---|---|---|---|---|---|---|---|---|---|---|---|---|
| P | R | F1 | Acc | P | R | F1 | Acc | P | R | F1 | Acc | P | R | F1 | Acc |
| 0.96 | 0.94 | **0.95** | 0.94 | 0.96 | 0.93 | 0.94 | 0.94 | 0.73 | 0.55 | 0.63 | 0.60 | 0.81 | 0.79 | 0.80 | 0.79 |

[†] Uses patches sampled from THOR-High.
[‡] Uses entire full-resolution video input.

Table 4. Precision (P), Recall (R), F1-score (F1), and Accuracy (Acc) for THOR II and baseline methods.

*6.2.2 Comparison to Baselines.* Table 4 compares the performance of THOR II against baseline methods that operate on the full video. Compared to SwinIR, which classifies activities using a single full-frame image, our method achieves a 32% higher F1-score despite operating on a single cropped patch. Relative to TimeSformer, which relies on a temporal window of frames, THOR II achieves a 15% improvement in F1-score. Qwen2-VL, which shares the same architecture as our method but uses the entire video data, achieves a comparable F1-score of 94%. These results highlight the effectiveness of multi-modal foundation models over traditional vision-based architectures. In particular, our method matches the performance of full-data methods while utilizing only 3% of the original input, enabled by the fine-tuning strategy described in Section 5.2.

*6.2.3 Generalization.* We evaluate the generalization ability of our method by comparing the cosine similarity between the generated captions and human-provided narrations in the Ego4D dataset. Specifically, we compute the similarity between captions produced by THOR II and narrations from two human annotators (A1 and A2). As a baseline, we also calculate the similarity between the two annotators' narrations. As shown in Figure 9, the cosine similarity between A1 and A2 (0.59) is comparable to the similarity between A1 and THOR II (0.54), as well as A2 and THOR II (0.53), indicating that our method produces captions that are semantically aligned with human descriptions.

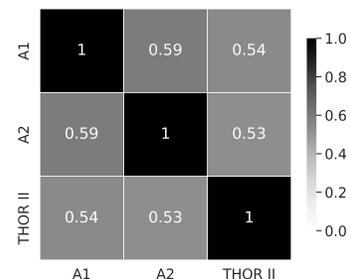

Fig. 9. Cosine similarity between narrations provided by annotators (A1 vs A2) and our method.

*6.2.4 Latency.* Figure 8 presents the latency profile of THOR II running on an iPhone 15 Pro. Figure 8-A compares the inference latency for processing full images versus hand-object patches. Using patches reduces input processing latency from 4.22 seconds to 1.21 seconds, representing a 71% decrease. This improvement is due to fewer visual tokens being processed, as shown in Figure 5. Figure 8-B shows the end-to-end latency, including caption generation. Since THOR II is fine-tuned to produce concise, hand-object-specific captions, total token generation time is shorter, reducing total latency from 6.63 seconds to 1.47 seconds per query, which corresponds to a 78% reduction.

*6.2.5 Power Consumption.* Figure 7-B shows the power consumption profile of model inference under THOR II compared to a baseline that uses the full-image input. In the baseline setup, 61% of the power is consumed during full-image processing. In contrast, our method processes only patches, reducing inference power usage to 24.2%. The shorter caption generation lowers power consumption even further by an additional 36%. Overall, our method reduces the model's total inference power requirement by 72.7%.

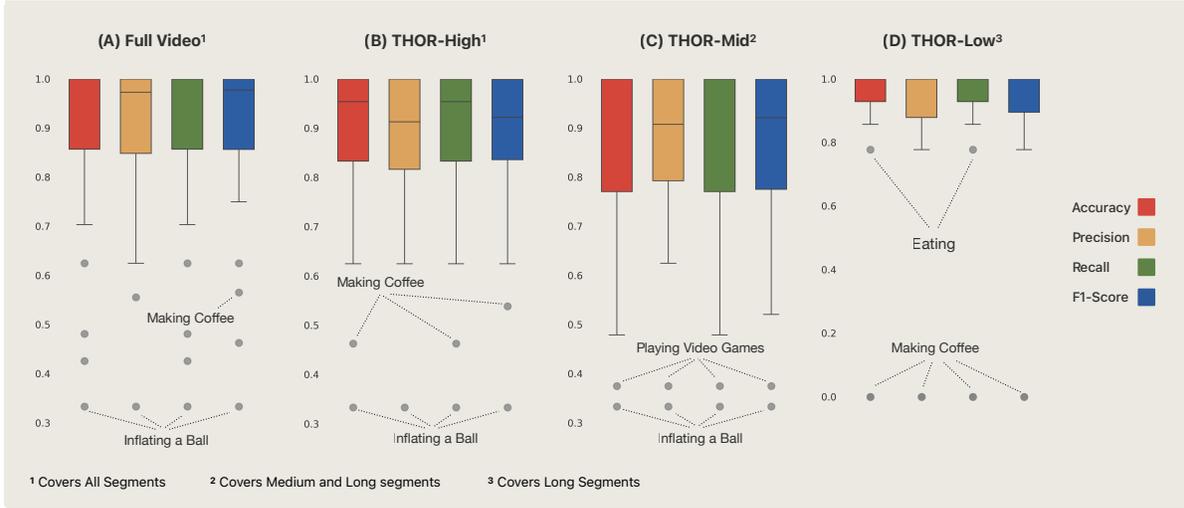

Fig. 10. Accuracy of different THOR variants in classifying 30 hand-related activities, compared to using the full video. Note that each variant is designed for segments of different lengths; therefore, their accuracies should not be directly compared to one another.

Table 3 isolates the cost of the our method inference under different input conditions. Providing hand-object patches as input consumes a modest 31 mWh, while generating text on those same patches adds an 18% overhead (36.5 mWh). In contrast, feeding the full-resolution image multiplies the energy draw by more than 2.5× (80 mWh), and generating output on full images pushes usage to 131 mWh—over four times compared to our method. This confirms that spatially constrained inputs cut energy by up to 76% relative to full-image inference.

## 7 Discussion and Future Work

### 7.1 Potential to Integrate Video Intelligence in Existing Systems: Wearable Camera Data Browser

Despite growing interest in using wearable cameras for health-related behavior monitoring, there has been limited effort toward the development of interactive tools that facilitate efficient browsing of wearable camera footage. Currently, dietitians investigating problematic eating behaviors, or researchers conducting ground truth annotation and visual confirmation, must manually review prolonged recordings that are both redundant and computationally burdensome. These practices diminish the scalability of such data-driven studies. Our method, THOR, addresses these inefficiencies by eliminating a significant portion of video frames that are not informative for hand-related activity recognition. This advancement opens the door to the creation of intelligent systems, such as an intelligent video browser tailored to the needs of clinical and research users. Unlike prior video browsing tools, which primarily rely on coarse visual cues such as hand presence or motion magnitude [21], our system extracts semantic descriptions from sampled patches using VLMs. These descriptions enable keyword-based search and filtering functionalities, allowing users to navigate video logs not by timestamps, but through meaningful textual summaries of the activities. By integrating THOR's spatio-temporal sampling and captioning pipeline into a browser interface, we envision a tool where users can query specific actions (e.g., "drinking from a mug" or "using phone while eating") and receive direct visual access to the most relevant segments. This approach dramatically reduces the cognitive load of reviewing video data and enhances the accessibility and utility of wearable camera recordings in behavioral health, diet monitoring, and activity analysis contexts.

## 7.2 Reshaping the Future of Activity Trackers

Conventional activity tracking applications typically rely on predefined taxonomies of activities for monitoring and classification. These taxonomies, while useful for population-level studies, constrain personalization and fail to accommodate user-specific behaviors that fall outside standardized categories. Leveraging the efficient spatio-temporal sampling pipeline of THOR, we demonstrated that lightweight vision-language models (VLMs) can operate in real-time on-device, generating semantic descriptions of hand-object interactions without requiring the full RGB video stream. A key advantage of using VLMs over conventional deep learning classifiers is their ability to produce free-vocabulary, natural language captions rather than relying on a fixed set of output classes. This capability introduces a new paradigm for activity tracking in which users are no longer limited to preset activity labels. We propose a custom activity tracker application that builds on this principle. Users can specify the activity they wish to monitor—ranging from broad behavioral categories (*e.g.*, screen time) to highly specific actions (*e.g.*, typing on a laptop while eating) and the system can detect and log instances of the specified behavior based on keyword-matching or semantic similarity with the generated captions. This approach enables a user-centered model of activity recognition that supports individualized goals and contexts, such as monitoring time spent on particular devices, habits, or multitasking behaviors.

## 7.3 Lowering the Barrier to Foundation Model Research

Recent advancements in foundation models, particularly VLMs have opened new frontiers in visual understanding, yet their computational demands present a significant barrier to entry. Fine-tuning and inference with these models often require high-end GPUs or access to distributed cloud computing infrastructure, resources that are typically available only to large academic labs or industry partners. This restricts broader participation in foundation model research and hinders the democratization of innovation. In this work, by embracing the principle of least privilege we demonstrate that it is possible to use the capabilities of large-scale VLMs without overwhelming infrastructure requirements. All experiments presented in this work were conducted using four GeForce RTX 2080 GPU, a consumer-grade device released in 2018, and without reliance on commercial cloud computing services. This illustrates a compelling path forward for smaller research labs, independent investigators, and educational settings. By promoting efficiency and accessibility, our work encourages a more inclusive foundation model research ecosystem, where scientific progress is not constrained by the availability of large-scale compute.

## 7.4 Assessment of Misclassified Activities

We show that our method achieves a high F1-score in detecting 30 different fine- and coarse-grained activities. However, Figure 10 shows that certain activities exhibit low accuracy across all variants of THOR as well as baseline methods. Below, we discuss possible reasons behind some of these failure cases and potential solutions.

*7.4.1 Underrepresented Objects.* We observed failure modes when participants were "inflating a ball" using a portable air pump. The VLM often mislabes the pump as a vacuum cleaner–due to similar handle geometry–or defaults to describing it as a generic "black object." We find that the model frequently misidentifies the air pump as a vacuum cleaner, likely due to the visual similarity in their handles, or simply describes it as a generic black object. Occasionally the model ignores the pump entirely andinstead focuses on the more familiar object (*e.g.,* the ball) and generates captions such as "playing with a ball." These errors may stem from the model's lack of exposure to objects like portable air pumps during pretraining. One way to address this limitation is targeted fine-tuning of the VLM on hand-object interactions involving underrepresented tools, enabling more precise recognition and reasoning in such contexts.

*7.4.2 Complex Multi-step Activities.* The activity "making coffee" presents a challenge for the model due to its procedural complexity and variation in execution. This activity typically involves a series of sub-actions, such as opening a coffee pack, filling water, grinding beans, boiling water, and choosing among brewing methods–each involving distinct tools, hand movements, and visual contexts. Such activities that span multiple sub-actions makes it difficult for the model to form a consistent representation of the overall activity. Additionally, participant-specific equipment (e.g., French press, pour-over, moka pot) introduces variability that fragments the model's representation. Consequently, the model may misclassify based on the most visually prominent step. Addressing this limitation may require training with more structured, multi-step activity annotations or incorporating temporal reasoning over longer sequences.

*7.4.3 Keyword Overlap and Semantic Ambiguity.* The activity "playing video games" often involves a gaming controller, which can serve as a distinctive visual cue. However, this activity overlaps with others that share similar contexts, such as "using computer." In many cases, a person using a computer may also be playing a game, and our keyword set includes terms like "playing games" and "using computer" for both situations. This overlap introduces ambiguity and can lead the model to misclassify one activity as the other. One possible way to address this issue is to move beyond exact keyword matching and instead compute semantic similarity between the generated captions and activity labels (see Section 5.3.3). This approach allows the system to recognize activities based on the overall meaning of the description, rather than relying on specific predefined terms, which can reduce confusion in overlapping scenarios.

*7.4.4 Concurrent Activities.* We observe several misclassifications in cases where participants perform multiple activities at the same time, or what some may characterize as secondary activities [55]. For example, in the "eating while using a phone" scenario, the model often generates captions that describe only one of the actions, such as "using a phone" or "eating," depending on which object, like a spoon or a phone, is more visually prominent in the patch. This limitation may be due to the training data, where each caption typically corresponds to a single dominant activity. As a result, THOR II can miss secondary actions in multi-tasking situations. One possible way to improve recognition in these cases is to incorporate multi-label or compositional annotations that reflect the presence of multiple hand-object interactions. Additionally, prompting strategies could be modified to encourage the model to enumerate all visible actions (e.g., "What are all the activities happening?"), rather than identifying just the most salient one. Finally, identifying the active hand, such as the one currently manipulating an object, may help determine which task the person is focused on, leading to more accurate activity descriptions.

## 7.5 Limitations and Future Work

*7.5.1 Activity-Boundary Segmentation.* In this work, we primarily focused on classifying activity segments and detecting their onset. In future work, we will evaluate and extend THOR to accurately delineate both start and end boundaries of each activity, enabling precise temporal segmentation of both fine- and coarse-grained activities.

*7.5.2 Device's Form Factor.* The current head-mounted prototype prioritizes sensor integration and data fidelity over user comfort. We anticipate that existing and emerging wearable platforms—such as smart glasses equipped with embedded thermal modules and compact RGB sensors—can seamlessly house our dual thermal-RGB camera system without compromising ergonomic design, or aesthetic appeal.

*7.5.3 Deployment.* Although our study with 14 participants was conducted in an uncontrolled setting, all analysis was performed offline. Several existing clinical trials and studies are incorporating wearable thermal and RGB cameras to study health-risk behaviors. Next steps include real-time THOR deployment in such large cohorts to evaluate system robustness, latency, and usability in long-term health and behavior monitoring studies.

## 8 Conclusion

We presented THOR, a dual-modality RGB sampling framework that adaptively modulates both when (temporal) and where (spatial) frames are captured. By leveraging thermal cues to identify transitions between hand-related activities and dynamically adjust the RGB sampling rate and spatially crops to the hand-object region, yielding a 95% F1-score on hand-activity recognition. By reducing RGB data usage, THOR addresses privacy concerns, lowers power consumption, reduces inference latency, and decreases storage requirements, all while matching the performance of full-video approaches. THOR thus provides a practical and scalable solution for longitudinal wearable-camera-based studies, particularly in applications related to health and behavior monitoring.

## A  Appendix

| Activity | Keywords |
|---|---|
| Using Phone | texting, smartphone, phone, watching a video, social media, taking a picture, taking a photo, sending a message, writing a message, texting |
| Assembling Puzzle | assembling puzzle, jigsaw puzzle, matching pieces, puzzle solving, puzzle |
| Cleaning Window Screen | window cleaning, squeegee, wipe screen, dust removal, clean window, glass, cleaning a window |
| Cutting Food | cutting board, knife, slice, chop, dicing, cut ingredients, cutting potato, cutting onion |
| Cutting Paper | cutting paper, scissors |
| Drawing | drawing, sketching, pen, pencil, illustration, color pencil |
| Eating | eating, eat, meal, dining, snacking, holding a spoon, holding a fork, making a sandwich |
| Folding Clothes | folding clothes, laundry, fold shirts, fold pants, fold towels, organize clothes, holding clothes, sorting clothes, sorting laundry |
| Grating Food | grating, grate cheese, shredding, grater, grate vegetables |
| Inflating | inflating, air pump, inflate ball, pump air, inflate tires |
| Ironing Clothes | ironing clothes, iron, press, steam iron |
| Making Coffee | making coffee, brew coffee, espresso, coffee machine, drip brew, coffee preparation, coffee, grinding coffee |
| Making Eggs | making eggs, fry eggs, scramble eggs, cook eggs, omelette, pan, cooking, holding an egg, cracking an egg, cracking eggs, stirring eggs, |
| Mopping Floor | mopping floor, mop, cleaning the floor |
| Peeling Fruit/Vegetable | peeling, peel fruit, peel vegetable, peeler |
| Playing Boardgame | board game, boardgame, monopoly, chess, checkers, dice |
| Playing Cards | card game, cards, poker, shuffle, playing cards |
| Playing Music | instrument, guitar, piano, violin, kalimba, marimba |
| Playing Video Games | video games, gaming, controller, console, playstation, xbox, nintendo, playing a video game |
| Reading a Book | reading, read book, book |
| Sewing | sewing, needle, thread, stitch, fabric, hemming, threads |
| Using Hammer | using hammer, hammer, nails, hammering, hammer |
| Using Screwdriver | screwdriver, screws, tighten, unscrew, turn screw |
| Using Blender | blending, blender, smoothie, mix ingredients |
| Using Computer | using computer, typing, coding, browsing, web browsing, laptop, using a computer, playing a video game, playing a game, using a mouse, working on a computer, holding a keyboard |
| Vacuuming | vacuuming, vacuum cleaner, clean floor |
| Washing Dishes | washing dishes, dish soap, scrub dishes, clean dishes, plate, bowl, sink, washing a pot, washing a mesh strainer, washing a pan |
| Washing Hands | washing hands, soap |
| Wiping Countertop | wiping countertop, wipe surface, clean counter, wiping |
| Writing | writing, write, pen, pencil, handwriting, notes, journal |

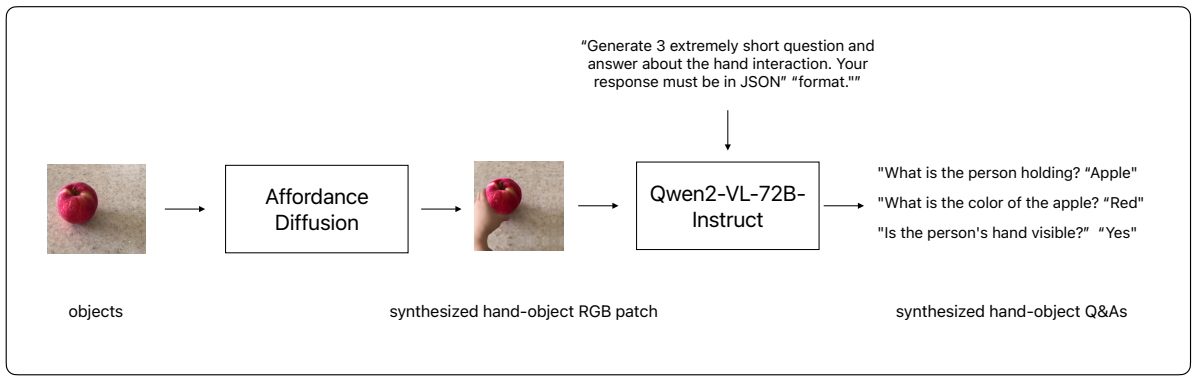

Fig. 11. Overview of our synthetic data generation pipeline. We feed the image of common objects into a diffusion model to synthesize hand-object interaction. Then we generate question and answer pairs using a larger-scale vision-language model to create a synthetic dataset for THOR II.